# Many-Objective Estimation of Distribution Optimization Algorithm Based on WGAN-GP

Zhenyu Liang, Yunfan Li, Zhongwei Wan

*Abstract*— Estimation of distribution algorithms (EDA) are stochastic optimization algorithms. EDA establishes a probability model to describe the distribution of solution from the perspective of population macroscopically by statistical learning method, and then randomly samples the probability model to generate a new population. EDA can better solve multi-objective optimal problems (MOPs). However, the performance of EDA decreases in solving many-objective optimal problems (MaOPs), which contains more than three objectives. Reference Vector Guided Evolutionary Algorithm (RVEA), based on the EDA framework, can better solve MaOPs. In our paper, we use the framework of RVEA. However, we generate the new population by Wasserstein Generative Adversarial Networks-Gradient Penalty (WGAN-GP) instead of using crossover and mutation. WGAN-GP have advantages of fast convergence, good stability and high sample quality. WGAN-GP learn the mapping relationship from standard normal distribution to given data set distribution based on a given data set subject to the same distribution. It can quickly generate populations with high diversity and good convergence. To measure the performance, RM-MEDA, MOPSO and NSGA-II are selected to perform comparison experiments over DTLZ and LSMOP test suites with 3-, 5-, 8-, 10- and 15-objective.

## I. INTRODUCTION

MULTIOBJETIVE optimal problems (MOPs) are common optimal problems in real world, and there already are several powerful estimation distribution algorithms (EDA) can deal with some optimal problems. But those algorithms like RM-MEDA [2] have good result only on such optimal problems that have objectives less than 3. In our project, we try to improve EDA to make it powerful on MaOPs.

### A. Multi-objective and Many-Objective

Since in real world, problems always depend on many factors, this is why we try to find a algorithm to deal with MOPs.

The definition of MOPs can be represented by this format:

$$\min y = F(\mathbf{x}) = (f_1(\mathbf{x}), f_2(\mathbf{x}), ..., f_M(\mathbf{x}))$$
$$s.t. \quad \mathbf{x} \in X \qquad (1)$$

where $X \in R^n$ is the decision space on Euclidean space with $\mathbf{x} = (x_1, x_2, x_3, ..., x_n) \in X$ is a vector and M is the number of objectives on problem $F$. For MOPs, M always is 2 or 3, and for those problems have objectives more than 3, called many-objective problems (MaOPs).

### B. Estimation Distribution Algorithm

In fact, EDA is a stochastic optimization algorithm. It has population which is a set of several solution for MOPs and offspring which is new solution created by algorithm and selection which is used to select better solution. But the main difference between EDA and normal EA is that EDA use a possibility model to create offspring and optimize model and population. The common framework of EDA is shown below [3]:

**Step 0) Initialization:** Set generation t as 0. Generate an initial population $P(0)$. Use objective function to evaluate each individual $x$ and set these vectors as $F(0)$. Create an initial model as $M(0)$.

**Step 1) Stopping Condition:** If stopping condition is met, stop and return population $P(t)$. Then use $f$ to calculate $F(t)$, and $F(t)$ is the set that approach to Pareto front.

**Step 2) Sampling:** Use model $M(t)$ to generate new offspring $Q(t)$ and combine population $P(t)$ and offspring $Q(t)$ as a new large population $P(t)$.

**Step 3) Selecting:** Use selecting method to get better individual in population that approach to Pareto front and set these better individuals as new population $P(t+1)$.

**Step 4) Updating:** Update model by $P(t+1)$ and $M(t)$, then set new model as $M(t+1)$.

**Step 5) Setting generation:** Set generation t as t+1 and go to **Step 1)**.

## II. BACKGROUND

### A. WGAN-GP

WGAN-GP [4] is an improved WGAN [3], it also use the thought of generative adversarial nets [5]. It sets input data as real data and to training generator and discriminator. Generator is used to approach noise to real data, and discriminator is used to evaluate the correctness of data that come from generator. Every time discriminator have evaluate the correctness of data, it will send feedback to itself and generator, generator will try to cheat discriminator and discriminator want to evaluate data correctly. In this process two net will improve their performance until termination. And the trained generator can be used to as a model to generate new offspring in EDA. The training process is shown in Fig.1.

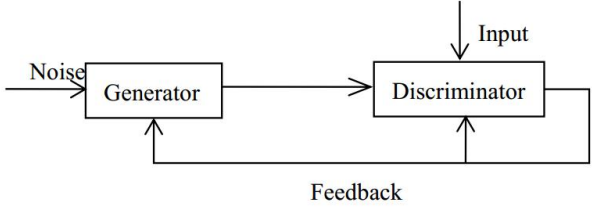

Fig. 1. WGAN-GP

The main improvement in WGAN-GP is that it used gradient penalty to improve loss function. In WGAN, though it use Wasserstein distance to calculate loss function, WGAN still have serious problem: loss function always tends to extreme value of constraint [4]. It cause WGAN can't be trained enough and seriously affect its performance. But With gradient penalty, it can smoothing weight of loss function, avoid the extreme situation.

*A. RVEA*

In RVEA, it uses reference vectors to select new population.

Reference vectors is unit vectors uniformly distributed inside the first quadrant [1]. It generated by norm. It can use a method called canonical simplex-lattice design [6] to create uniformly distributed point inside first quadrant, then use its norm can get uniformly distributed reference vectors. These reference vectors will refer to Pareto front later.

In selection part, there are three steps need to do. First, to translate objective value of individuals to first quadrant. Then divide those individuals into N partition, where N is the number of reference vectors. Lastly, use Angle-Penalized Distance (APD) Calculation [1] to calculate the APD between reference vectors and select the individual have minimal APD on its translated objective value for each reference vectors.

After selection, it is necessary to adapt reference vectors to make sure its reference function.

### III. PROPOSED ALGORITHM

*A. Main Framework*

The main framework of our algorithm Reference Vector Guided Evolutionary Algorithm Based On WGAN-GP(RVEA-WG) is shown below[1]:

**Step 0) Initialization:** Generate the initial population $P_0$ with $N$ randomized individuals and a set of
unit reference vector $V_0 = \{\mathbf{v}_{0,1}, \mathbf{v}_{0,2},...,\mathbf{v}_{0,N}\}$.
**Step 1) Stopping Condition:** If stopping condition is met, stop and return population $P_t$. Then use $f$ to calculate $F_t$.
**Step 2) Offspring Creation:** Use WGAN-GP to create the new offspring.
**Step 3) Mutation:** The offspring perform the polynomial mutation operation and then combine with the parent population to form a new population.
**Step 4) Selecting:** Use reference vector-guided selection to select N individuals.

**Step 5) Adaptation:** Adapt the reference vector.
**Step 6) Setting generation:** Set generation t as t+1 and go to **Step 1)**.

*B. Offspring Creation*

In our paper, we generate offspring by WGAN-GP[4]. We use 3 Fully connected layer as network, and a tanh function is used to constraint range of output. Every generation we will use two methods in WGAN-GP. One is pre-train which only trains our discriminator D, and another train
part which trains whole network.

In pre-train part, we set better data as real data and set others as bad data, and train discriminator D that make it more close to real data and more far from bad data. And those data are from our last selection. We input selected data as better data and input eliminated data as bad data., which let this pre-train more like nature selection. The process of pre-train is shown in Fig.2.

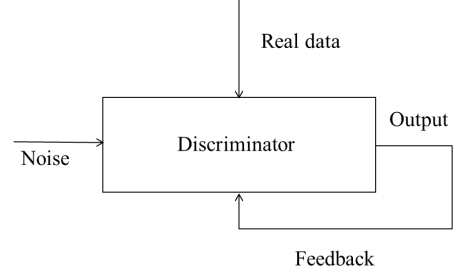

Fig. 2. Pre-train

The training process show in the Fig.1. There are Generator G and Discriminator D in WGAN-GP. Firstly, generate some noise data by standard normal distribution model or uniform distribution model. The noise data is the input of G and put the output into D. Secondly, create some good Pareto set as real data set $p_r$ and put those data into our WGAN-GP to train a generator G that can transfer standard normal distribution to Pareto set. With WGAN-GP training, we will get results that are increasingly similar to real data. Also, the data are smooth and there is no gradient explosion or gradient disappearance.

*C. Reference Vector-Guided Selection[1]*

There are four steps in reference vector-guided selection: 1) objective value translation; 2) population partition; 3) APD calculation; and 4) the elitism selection.

1) *Objective Value Translation*: $F_t = \{\mathbf{f}_{t,1}, \mathbf{f}_{t,2}, ...,\mathbf{f}_{t,|P_t|}\}$ donates the objective values of the population. Objective values translate from $F_t$ to $F_t'$ via

$$\mathbf{f}_{t,i}' = \mathbf{f}_{t,i} - \mathbf{z}_t^{min} \qquad (2)$$

, where i = 1, … , $|P_t|$, $\mathbf{f}_{t,i}$ and $\mathbf{f}_{t,i}'$ are the objective vectors of individual i before and after the translation. Also $\mathbf{z}_t^{min} = (z_{t,1}^{min}, z_{t,2}^{min},..., z_{t,m}^{min})$ represents the minimal objective values calculated from $F_t$.

2) *Population Partition*:
After the translation, the population will be partitioned into N subpopulation by associating each individual with its closest reference vector. To get the closest reference vector, we can

calculate the angle between reference vector and objective vector:

$$cos\theta_{t,i,j} = \frac{\mathbf{f}'_{t,i} \cdot \mathbf{v}_{t,j}}{\|\mathbf{f}'_{t,i}\|} \quad (3)$$

where $\theta_{t,i,j}$ represents the angle between objective vector $\mathbf{f}'_{t,i}$ and reference vector $\mathbf{v}_{t,j}$.

Then choose the maximal cosine to partition:

$$\bar{P}_{t,k} = \{I_{t,i} | k = argmax \, cos\theta_{t,i,j}\} \quad (4)$$

where $I_{t,i}$ denotes the $i$th individual in $P$t.

3) *Angle-Penalized Distance calculation*: Calculate APD by:

$$d_{t,i,j} = (1 + P(\theta_{t,i,j})) \cdot \|\mathbf{f}'_{t,i}\| \quad (5)$$

where $P(\theta_{t,i,j})$ is a penalty function related to $\theta_{t,i,j}$.

$$P(\theta_{t,i,j}) = M \cdot \left(\frac{t}{t_{max}}\right)^\alpha \cdot \frac{\theta_{t,i,j}}{\gamma_{v_{t,j}}} \quad (6)$$

$$\gamma_{v_{t,j}} = \min_{i \in \{i,...N\}, i \neq j} \langle v_{t,i}, v_{t,j} \rangle \quad (7)$$

where M is the number of objectives, N is the number of reference vectors, $t_{max}$ is the predefined maximal number of generations, $\gamma_{v_{t,j}}$ is the smallest angle value between reference vector and the other reference vectors in the current generation, and α is a user defined parameter controlling the rate of change of $P(\theta_{t,i,j})$.

4) *Elitism Selection*:
Select the smallest APD of population as the elite population.

### D. Reference Vector Adaptation[1]

Instead of normalizing the objectives, the algorithm adapt the reference vectors according to the ranges of the objective values in the following manner:

$$\mathbf{v}_{t+1,i} = \frac{\mathbf{v}_{0,i} \circ (\mathbf{z}^{max}_{t+1} - \mathbf{z}^{min}_{t+1})}{\|\mathbf{v}_{0,i} \circ (\mathbf{z}^{max}_{t+1} - \mathbf{z}^{min}_{t+1})\|} \quad (8)$$

$\mathbf{v}_{t+1,i}$ denotes the $i$th adapted reference vector for the next generation $t+1$, $\mathbf{v}_{0,i}$ denotes the $i$th uniformly distributed reference vector of initial stage and $\mathbf{z}^{max}_{t+1}$ and $\mathbf{z}^{min}_{t+1}$ denote the maximum and minimum values of each objective function in the $t+1$ generation.

## IV. COMPARATIVE STUDIES

In this part of the experiment, we will use DTLZ[8] and LSMOP test suites to evaluate the performance of the four algorithms RVEA-WG, RM-MEDA[7],MOPSO[9]and NSGA-II[10]. After that, we will get the IGD value from the result of the algorithm. The specific process is as follows. The brief differences between the two test suites are as follows:

### A. DTLZ

DTLZ is a scalable test problems for evolutionary multi-objective optimization and named with the first letter of the last names of the authors. In DTLZ, all problems are n-dimensional continuous mutl-objective problems.

### B. LSMOP

LSMOP is test problem suite for large scale multi-objective and many-objective optimization , it is used for Large-scale benchmark MOP.

Run RVEA-WG, RM-MEDA, MOPSO and NSGA-II in 15 generations and in RVEA-WG the epoch of network is 40. Compare four conditions, the first one is that population size is 105, objective value is 3 and the number of the variable is related to objective value. The second one is that population size is 132, objective value is 6 and the number of the variable is related to objective value. The third one is that population size is 156, objective value is 8 and the number of the variable is related to objective value. The forth one is that population size is 275, objective value is 10 and the number of the variable +is related to objective value.

TABLE I are the IGD of three LSMOP running the four algorithm. TABLE II are the IGD of four DTLZ running the four algorithm. We run each problem 10 times and calculate the mean IGD value. The bolded number is the minimum value per line. M is objective value and related population size is shown above.

| Problem | M | RVEA-WG | RM-MEDA | MOPSO | NSGA-II |
|---|---|---|---|---|---|
| LSMOP1 | 3 | **8.9660e-1** | 9.3497e+0 | 2.8367e+0 | 1.0462e+1 |
| | 6 | **9.5540e-1** | 1.0366e+1 | 2.1052e+0 | 1.1892e+1 |
| | 8 | **9.8940e-1** | 1.0242e+1 | 2.4351e+0 | 9.3969e+0 |
| | 10 | **9.8060e-1** | 1.0845e+1 | 2.2032e+0 | 9.3989e+0 |
| LSMOP2 | 3 | 3.617oe-1 | 1.0600e-1 | 8.2040e-2 | **1.0339e-1** |
| | 6 | 3.8610e-1 | 2.4165e-1 | 3.1778e-1 | **2.1282e-1** |
| | 8 | 4.3770e-1 | 2.8356e-1 | 4.1125e-1 | **2.4900e-1** |
| | 10 | 4.2710e-1 | 2.9253e-1 | 4.2416e-1 | **2.3517e-1** |
| LSMOP3 | 3 | **4.9870e+0** | 1.7112e+1 | 1.5528e+1 | 2.0738e+1 |
| | 6 | 5.0498e+0 | **2.4165e-1** | 3.6015e+3 | 3.8659e+1 |
| | 8 | **3.7771e+0** | 2.6565e+1 | 3.7643e+3 | 4.0888e+1 |

| | 10 | **2.0701e+0** | 2.7312e+1 | 3.0435e+3 | 3.9649e+1 |

TABLE I. IGD value on different LSMOP problems with objective value M = 3, 6, 8, 10

| Problem | M | RVEA-WG | RM-MEDA | MOPSO | NSGA-II |
|---|---|---|---|---|---|
| DTLZ1 | 3 | 2.3646e+1 | 2.7825e+1 | 2.7429e+1 | **1.0579e+1** |
| | 6 | 1.9929e+1 | 2.1439e+1 | 7.5789e+0 | **5.0755e+1** |
| | 8 | **1.6419e+1** | 2.7696e+1 | 2.0159e+1 | 8.601e+1 |
| | 10 | **1.7714e+1** | 2.3657e+1 | 2.4293e+1 | 7.1503e+1 |
| DTLZ2 | 3 | 3.9960e-1 | 1.4607e-1 | **1.1059e-1** | 9.2858e-2 |
| | 6 | 7.3810e-1 | 9.1624e-1 | **5.2301e-1** | 7.6691e-1 |
| | 8 | 8.6080e-1 | 1.1633e+0 | **7.2121e-1** | 1.0489e+0 |
| | 10 | 9.5500e-1 | 1.3075e+0 | **7.5562e-1** | 1.1540e+0 |
| DTLZ3 | 3 | 2.8304e+2 | 2.3271e+2 | 3.6043e+2 | **1.3438e+2** |
| | 6 | 2.7882e+2 | **2.5872e+2** | 3.6328e+2 | 6.4573e+2 |
| | 8 | **2.9148e+2** | 3.9882e+2 | 4.2759e+2 | 7.4846e+2 |
| | 10 | **2.6022e+2** | 2.6548e+2 | 3.7811e+2 | 5.5384e+2 |
| DTLZ4 | 3 | 8.3780e-1 | 3.8680e-1 | **2.7209e-1** | 5.5776e-1 |
| | 6 | 1.0731e+0 | 7.6590e-1 | **6.9616e-1** | 7.3058e-1 |
| | 8 | 1.1263e+0 | 9.6043e-1 | **7.5503e-1** | 8.9825e-1 |
| | 10 | 1.1196e+0 | 1.0866e+0 | **7.6855e-1** | 1.0186e+0 |

TABLE II. IGD value on different DTLZ problems with objective value M = 3, 6, 8, 10

## V. CONCLUSION

According to the tables, we can find that RVEA-WG have a good performance in LSMOP and MOPSO have a good performance in DTLZ. With higher objective value, RVEA-WG have better performance. However, Our algorithm still has room for improvement. The first is the network. Our network can also be tuned better to get better populations. The second is the selection part. All in all, we will continue to optimize every part of the algorithm.

REFERENCES

[1] R. Cheng, Y. Jin, M. Olhofer, B. Sendhoff, "A Reference Vector Guided Evolutionary Algorithm for Many-Objective Optimization", IEEE Trans. Evol. Comput., vol. 20, no. 5, Oct. 2016
[2] Q. Zhang, A. Zhou and Y. Jin, "RM-MEDA: A Regularity Model-Based Multiobjective Estimation of Distribution Algorithm", IEEE Trans. Evo.Comput., vol. 12, no. 1, February 2008
[3] M. Arjovsky, S. Chintala, and L. Bottou1, "Wasserstein GAN", arXiv:1701.07875v3 [stat.ML] 6 Dec 2017
[4] I. Gulrajani, F. Ahmed, M. Arjovsky, V. Dumoulin, A. C. Courville, "Improved Training of Wasserstein GANs", in neural information processing systems, NIPS, 2017, pp.5767-5777.
[5] I. J. Goodfellow, J. Pougetabadiey , M. Mirza, B. Xu, D. Wardefarley, S. Ozairz , A. Courville, Y. Bengio, "Generative Adversarial Nets ", in neural information processing systems, NIPS, 2014, pp.2672-2680.
[6] J. A. Cornell, Experiments With Mixtures: Designs, Models, and the Analysis of Mixture Data. Hoboken, NJ, USA: Wiley, 2011.
[7] Q. Zhang, A. Zhou, and Y. Jin, RM-MEDA: A regularity model-based multiobjective estimation of distribution algorithm, IEEE Transactions on Evolutionary Computation, 2008, 12(1): 41-63.
[8] Laumanns M, Thiele L, Deb K, and Zitzler E, Combining Convergence and Diversity in Evolutionary Multiobjective Optimization. Evolutionary Computation, 10(3): 2002; 263-282.
[9] C. A. Coello Coello and M. S. Lechuga, MOPSO: A proposal for multiple objective particle swarm optimization, Proceedings of the IEEE Congress on Evolutionary Computation, 2002, 1051-1056.
[10] K. Deb, A. Pratap, S. Agarwal, and T. Meyarivan, A fast and elitist multiobjective genetic algorithm: NSGA-II, IEEE Transactions on Evolutionary Computation, 2002, 6(2): 182-197.